\documentclass[runningheads]{llncs}
\usepackage{graphicx}

\usepackage{amssymb}
\usepackage[normalem]{ulem}
\usepackage{amsmath}

\usepackage{subcaption}
\usepackage{color}
\newcommand{\roberto}[1]{\textcolor{blue}{{\it [Rob says: #1]}}}
\newcommand{\francesco}[1]{\textcolor{red}{{\it [Francesco says: #1]}}}

\newcommand{\commento}[1]{}

\begin{document}
\title{A general approach to compute the relevance of middle-level input features}
%\inst{1}\inst{2,3}
\author{Andrea Apicella \and
Salvatore Giugliano \and
Francesco Isgr\`o \and
Roberto Prevete
}
\authorrunning{A. Apicella et al.}
\institute
{Dipartimento di Ingegneria Elettrica e delle Teconologie dell'Informazione \\
Universit\`a degli Studi di Napoli Federico II, Italy
\email{}\\
\url{}
}
\maketitle              % typeset the header of the contribution
\begin{abstract}
This work proposes a novel general framework, in the context of eXplainable Artificial Intelligence (XAI), to construct explanations for the behaviour of Machine Learning (ML) models in terms of middle-level features which represent perceptually salient input parts. One can isolate two different ways to provide explanations in the context of XAI: low and middle-level explanations. Middle-level explanations have been introduced for alleviating some deficiencies of low-level explanations such as, in the context of image classification, the fact that human users are left with a significant interpretive burden: starting from low-level explanations, one has to identify properties of the overall input that are perceptually salient for the human visual system.
However, a general approach to correctly evaluate the elements of middle-level explanations with respect ML model responses has never been proposed in the literature. 

We experimentally evaluate the proposed approach to explain the decisions made by an Imagenet pre-trained VGG16 model on STL-10 images and by a customised model trained on the JAFFE dataset, using two different computational definitions of middle-level features and compare it with two different XAI middle-level methods. 
The results show that our approach can be used successfully in different computational definitions of middle-level explanations.
\keywords{XAI  \and Machine Learning \and Middle-level features.}
\end{abstract}
%

%\francesco{A volte si sono usate espressioni del tip \emph{image to explain}, che non è corretto, in quanto non spieghiamo l'immagine, cosa che fa in un certo senso il classificatore. Credo di averli aggiustati tutti, ma potrebbe essermene sfuggito qualcuno}

\section{Introduction}
\label{sec:introduction}

In the last years, Machine Learning (ML) approaches have been widely used to address several challenges in Artificial Intelligence (AI), such as image \cite{springenberg2015} and text classification \cite{devlin2018} problems, multi-target regression \cite{reyes2019} and robot navigation \cite{richter2018}. 
However, a large part of these approaches suffers from a pervasive lack of transparency also connected to the problem of explaining their behaviour in terms that are easy to understand for  human beings \cite{letham2015}.  
Indeed, it seems that the better ML systems become in terms of their performance, the harder it is to understand the underlying mechanisms and explain their behaviours \cite{adadi2018}. 
For this reason, ML systems are often considered as black-box systems \cite{adadi2018} insofar as their decisions are hard to interpret in terms of meaningful input features. 
Thus, generating explanations for ML system behaviours that are \textit{understandable to human beings} is a central scientific and technological issue addressed by the rapidly growing AI research area of  eXplainable Artificial Intelligence (XAI). 

\commento{
\roberto{questa parte non so se toglierla: To illustrate, consider recent ML work in the healthcare domain which is aimed at developing medical diagnostic tools which identify diseases from bio-physical features or medical imaging inputs \cite{martinez2018convolutional,de2018clinically}. 
There, explanations of system outcomes would be useful to identify undesirable artefacts and biases in training sets, to build trust in their responses and to prescribe confidently related clinical treatments. }
\francesco{Togliamola pure. E' un WS su XAI.}
}

The literature counts various strategies to make ML systems - especially those endowed with Deep Neural Network (DNN) architectures \cite{montavon2017} - interpretable and explainable \cite{erhan2009,nguyen2016}. 
XAI approaches to the explanation problem can be classified in several ways according to which properties are taken into account \cite{adadi2018,guidotti2018,zhang2018,nguyen2019}. 
A key distinction is  between \textit{low-level} and \textit{middle-level} input feature approaches.
Low-level feature approaches to XAI attempt to explain the output of an ML system in terms of low-level features of the input such as pixels in case of image classification problems.
One of the most successful methods for this type of approaches is the Layer-wise Relevance Propagation (LRP) \cite{bach2015}, which associates a relevance value to each input element (to each pixel in the case of images) as an explanation of the ML model response.  
Thus, human users are left with a significant interpretive burden: starting from the relevance values of each input element (pixel), one has to identify properties of the overall input that are perceptually salient for the human visual system.
A method which attempt to alleviate this drawback of low-level approaches to explanation have been proposed in \cite{Apicella2019d,apicella2020middle}, where explanations are provided in terms of middle-level properties (\textit{atoms}) of the input which represent perceptually salient input parts \cite{barghout2015spatial}. 

A popular method which is also based on middle-level properties of the input is LIME \cite{ribeiro2016}, which returns a set of image parts (\textit{superpixels}), that could have driven the ML model to the given answer. 
This set of superpixels can be then considered as an explanation to the ML model response. 
To the best of our knowledge, these types of approaches can be classified as \textit{model-agnostic}.

Model-agnostic approaches correspond to XAI methods which are independent of the ML model to be explained  \cite{adadi2018}, i.e.,  model-agnostic solutions are built relying only the relation between ML model inputs and outputs, without any consideration about the ML model internal state. 
Although this property ensures the applicability of these approaches to any ML model, on the other hand, how we will discuss more in details in Section \ref{sec:method}, the explanations of the model-agnostic methods 
could not be fully related to the actual causal relationships between model's inputs and outputs which have contributed to the given model response.
%can be considered just a \textit{hypothesis} about the ``motivation behind'' the model's output, saying nothing about the real input-output relations which have contributed to the given model answer. 
For instance, LIME returns an explanation inspecting the behaviour of the model in the neighbourhood of the input, but nothing ensures that, for that particular input instance, the answer of the classifier has a totally different explanation (for example, a particular on the background of the specific input image which the model has already seen during the training stage, making the model biased). 

In this paper, we propose a new method, that we called Middle-Level Feature Relevance (MLFR),  based on a variation of LRP that, instead of returning a relevance value for each input pixel, returns relevance values for a given set of middle-level features. This method can be applied whenever a) the input of a ML system can be
encoded and decoded on the basis of middle-level features, and b) LRP can be applied on both the ML model and the decoder (see Section \ref{sec:method} for further details). 
In this sense we consider MLFR as a \textit{a general framework} insofar as it can be applied on several different computational definitions of middle-level features as we will discuss in Section \ref{sec:method}. 
Notice that MLFR is not a model-agnostic approach, however it can be applied to a large class of ML models as well as LRP \cite{bach2015}, for example feedforward neural networks architectures such as shallow network and deep networks.

This paper is structured as follows. Section \ref{sec:related} briefly reviews the related literature;  Section \ref{sec:method} describes the proposed architecture;  experiments and results are discussed in Section \ref{sec:exp}; the concluding Section \ref{sec:conclusion} summarises the main results of the proposed explanation framework and outlines some future developments.

\section{Related works}
\label{sec:related}
Many XAI methods have been proposed since explainability is now a sought for requirement for AI solution.  
The literature proposes several reviews trying to categorise/distinguish the existing methods  \cite{montavon2017,adadi2018,zhang2018,guidotti2018} looking at different properties of the XAI methods.
According to these categorisations, our method can be classified as a \textit{white-box} and \textit{local} XAI approach.  White-box approaches require access to the internal structure of the ML model  \cite{adadi2018}. 
By contrast, black-box, or \textit{model-agnostic}, approaches provide explanation methods which are independent of the ML model \cite{adadi2018}, i.e., they need access only to the input-output relations of the ML model. 
Local approaches provide explanations for each given input, while the goal of global approaches is to produce an explanation for the whole behaviour of the ML system \cite{montavon2017}.

Many model-agnostic approaches are based on \textit{proxy-models} \cite{craven1996,caccavale2019,oh2019} or some type of maximisation of the ML model response with respect to the input, such as the Activation-Maximisation (AM) method \cite{erhan2009}.  
Proxy models are models behaving similarly to the original model, but in a way that it is
easier to explain \cite{gilpin2018}. 
Approaches based on AM method enables one to determine the input that makes the output of the ML model as close as possible to the model's initial response, for example, in case of classification problems, given $C_k$ as the response of the ML model, one maximises the $P(C_k|\textbf{x})$ with respect to $\textbf{x}$ satisfying some constraints on $\textbf{x}$. 
Notice that the explanations of the model-agnostic methods suffer from the lack of information about the actual input-output causal relationships which have contributed to the given ML model answers; thus these explanations may not be related to the specific ML model response to be explained \cite{li2020survey}.

Another critical distinction is based on the granularity level of the explanations. 
In fact, several XAI solutions provide explanations in terms of low-level input features. 
For instance, in image classification problems the output of an ML system is explained considering low-level features of the input image in terms of salience maps where to each pixel is associated a relevance value which quantifies the degree of importance of that pixel to cause the ML model response. 
Among the approaches of this type, Layer-wise Relevance Propagation (LRP) \cite{bach2015}, is the  most popular in the literature. 
LRP is a white-box approach, although it applies to many ML models such as deep networks. 
Notice that it is a general framework rather than a specific method insofar as it is defined as a set of constraints that an XAI algorithm should satisfy. 
Thus, different XAI algorithms with different explanations may be appropriate under these constraints \cite{bach2015}. For example, Deep-Taylor Decomposition \cite{montavon2017} can be interpreted as a way of obtaining LRP. %\roberto{da verificare, in ogni caso si possono citare le diverse regole alfa-beta, zeta e cosi' via}.  

In this type of approaches, human users are left with a significant interpretive burden: starting from the relevance values of each input element (pixel), one has to identify properties of the overall input that are perceptually salient for the human visual system.
Thus, to alleviate this cognitive burden, an alternative model-agnostic method, called Explanation-Maximization (EM), was proposed in \cite{Apicella2019b,Apicella2019d,apicella2020middle}. 
EM, which also applies in different areas, was instantiated in the context of image classification systems. 
EM obtains sets of perceptually salient middle-level properties of input images by applying sparse dictionary learning techniques and a variant of AM. 
These middle-level properties are used as building blocks for explanations of image classifications. 
However, this approach suffers from the typical shortcomings of the model-agnostic ones as regards the reliability of the explanations given, as previously discussed.
Among other methods using middle-level input features to build explanations about ML model responses, LIME \cite{ribeiro2016} can be considered the most popular in the literature. 
It is model-agnostic and based on a proxy-model: it explains the output of an ML system by observing its behaviour on perturbations of its input. 
The input is partitioned in a collection of \textit{components} (super-pixel in the case of images); perturbed inputs are composed of specific superpositions of these components. 
Perturbed inputs and outputs are used to construct a local linear model which is used as a simplified proxy for the original ML system in the neighbourhood of the input.
Thus, from the proxy, it is possible to infer an explanation of the original ML model response.
However, the faithfulness of the proxy with respect to the original model remains an open issue \cite{li2020survey}. %\roberto{aggiungere altre referenze}.  
Other methods, based on LIME, as in  Ribeiro et al. \cite{ribeiro2018} and Guidotti et al. \cite{guidotti2018local},  return explanations in terms of decision rules that are used as local conditions for decisions.

The method we propose in this paper differs from the works mentioned above in the following aspects, as it can be seen as a general framework to obtain middle-level explanations analysing the actual input-output relationship defined by the ML model.
Thus, different definitions of middle-level input features with different resulting solutions may be possible under this general framework. 
The only constraints are that the input can be encoded and decoded based on the defined middle-level input features and that LRP can be applied on both the ML model to be explained and the input decoder.  

\section{Middle-Level Relevance}
\label{sec:method}
%The proposed approach relies on the LRP explanation method \cite{bach2015}. 

%\roberto{BEGIN MERGE ROB, ANDREA TENTATIVO (NOTAZIONE DA UNIFORMARE ed aggiustare'):}
Given an ML model $M$ which receives an input $\mathbf{x} \in R^d$ and outputs $\mathbf{y} \in R^c$,
let us suppose that $\mathbf{x} $ can be decomposed in a set of $m$ middle-level features
$\mathbf{v}^i$ each one encoded by a value $u_i$. 
More formally, we suppose that a decoder $D:(\mathbf{V},\mathbf{u}) \xrightarrow[]{} \mathbf{x} \in R^d$ exists. 
Where $\mathbf{V}=\{\mathbf{v}^i\}_{i=1}^m$ is the set of $\mathbf{v}$'s middle-level features 
and $\mathbf{u}\in R^m$ encodes $\mathbf{x}$ in terms of the middle-level features. 
For example, in an image classification problem, a possible set of middle-level features can be the result of a segmentation algorithm on the input image $\mathbf{x}$ which produces a partition of $\mathbf{x}$ in $m$ regions or partitions $\{P_i\}_{i=1}^m$. Each image's partition $P_i$ can be represented by a vector $\mathbf{v}^i \in R^d$ such that their summation is equal to $\mathbf{x}$, in this case the decoder is a linear combination of the $\mathbf{v}^i$ with all the coefficients equal to $1$, which represent the encoding of the image $\mathbf{x}$ on the basis of the $m$ partitions (see Section \ref{subsec:dec_by_segmentation}). 
%Then, if LRP is applicable on both $M$ and $D$, then we stack $M$ on the top of $D$, we obtain a new model $(D,M)$ %FINE ROB TENTATIVO: 
%BEGIN ANDREA TENTATIVO

\begin{figure}
    \centering
    %\hspace*{-.9in}
    \begin{subfigure}{.7\textwidth}
        \includegraphics[width=1\linewidth]{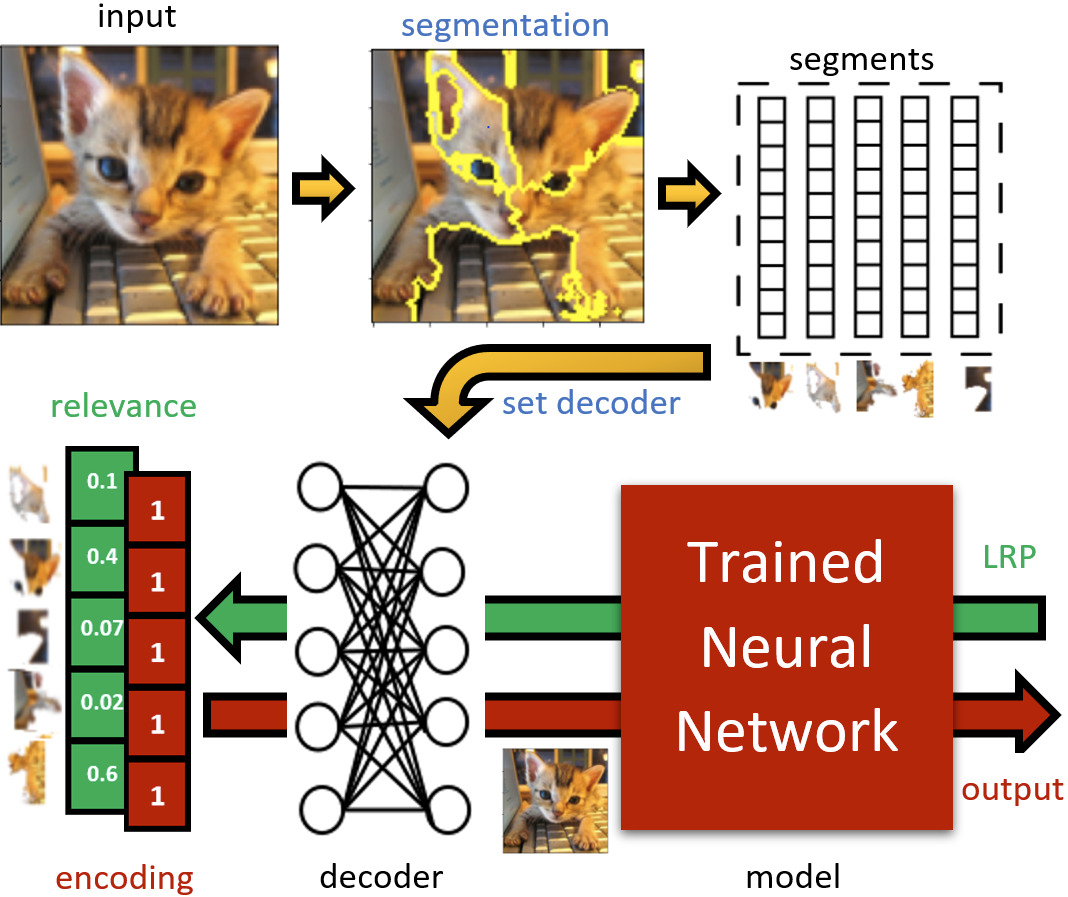}
        \caption{The segmentation-based approach.}
    \end{subfigure}
    
    \begin{subfigure}{.7\textwidth}
        \includegraphics[width=1\linewidth]{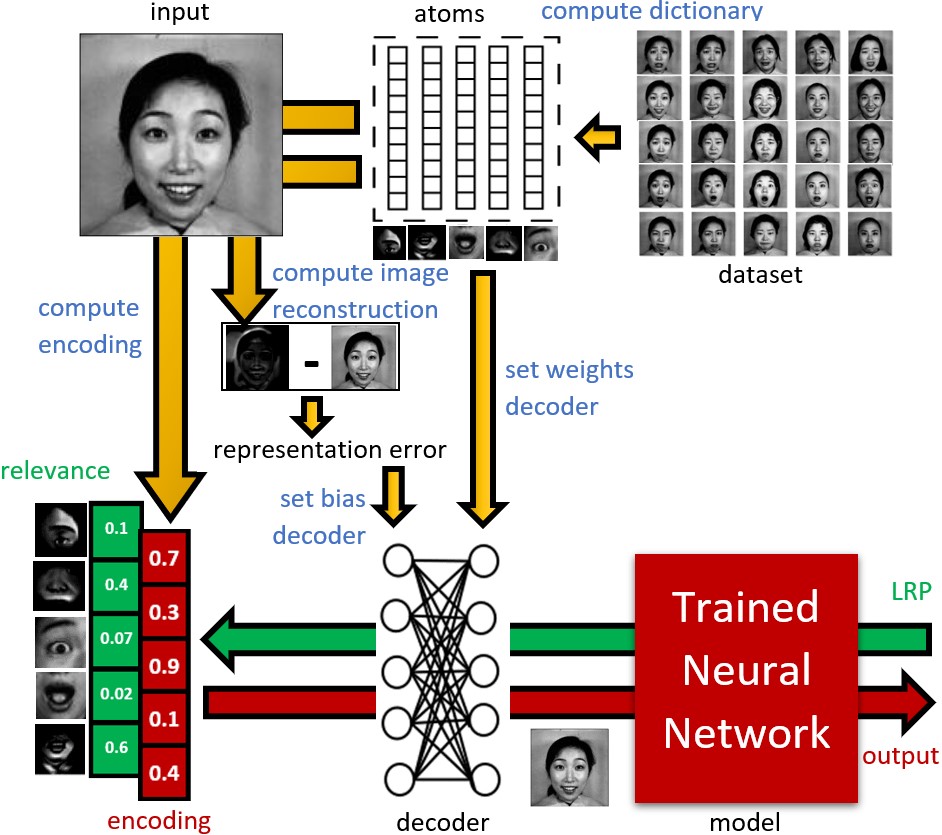}
        \caption{The dictionary-based approach.}
    \end{subfigure}
    \caption{A description of the proposed method (MLFR) using two different types of middle-level features. (a) After segmenting the input, the segments are used as weights for the decoder, so feeding the decoder with the $\mathbf{1}$s  is equivalent to give the input image to the trained neural network. After, the LRP algorithm is used to obtain the segment relevances  (see section \ref{subsec:dec_by_segmentation} for further details).
    (b) Having a dictionary, and an input encoding which best approximates the input image, we can use the dictionary and the representation error respectively as weights and bias of the decoder. So, feeding the decoder with the input encoding is equivalent to give the network the input image. After, the LRP algorithm is used to obtain the atom relevances  (see section \ref{subsec:dec_by_dictionary} for further details).}
    \label{figMethod}
\end{figure}

%\setlength{\parskip}{-0.0em} % sal: corregge spazi tra i paragrafi "anomali"
Then, if we can use LRP on both $M$ and $D$, we can apply it on the model $M$ and use the obtained relevance values to apply LRP on $D$ thus getting a relevance value for each middle-level feature.
In other words, we can stack $D$ on the top of $M$, thus obtaining a new model $DM$ which receives as input $\mathbf{u}$ and outputs $\mathbf{y}$, and uses LRP propagation on $DM$ from $\mathbf{y}$ to $\mathbf{u}$. 
Let us take as an example of $M$ a neural network composed of $L$ layers. 
The LRP procedure computes a set of relevance values for any given layer $l$ composed of $k_l$ neurons as the combination of the scores assigned to each neuron of $l$, representing the importance of each node for the network's output. 
The scores are computed by propagating the relevance values from the output layer to the input layer in a back-propagation fashion. 
Similarly, let us consider a shallow neural network composed of $m$ input values $u_i$, $d$ output neurons with biases equal to $0$, the identity as activation functions, and one hidden layer of weights $W=V$. 
This network can be seen as a decoder $D$ where the weights associated with the connections going from each input value $u_i$ to all output neurons represent the middle-level feature vector $\mathbf{v}^i$. 
If we stack the shallow network/decoder $D$ on the top of the $L$-layer model $M$, we obtain a new neural network model $DM$ composed of $L+1$ layers. 
Then, we can apply the LRP procedure on the whole $DM$ model and obtain relevance values which designate what input's middle-level features have most contributed on the outcome $y_{i}$ (see Figure \ref{figMethod}). 
In other words, we search for a relevance vector $\mathbf{r}\in \mathbb{R}^m$ which helps the user to know each middle-level feature of $\mathbf{x}$ how much has contributed to the ML model answer $y_{i}$. 
Note that, this approach can be generalised to any decoder to which LRP applies. 
For instance, we can consider any dictionary learning approach, as for example \cite{jenatton2010sspca,lee2001nmf,tessitore2011designing} (see Section \ref{subsec:dec_by_dictionary} for more details), where each input $\mathbf{x}$ can be decomposed as $\mathbf{x}=\mathbf{V} \mathbf{u} + \mathbf{\epsilon}$,   $V$ is a dictionary of middle-level features and $\mathbf{\epsilon}$ is the reconstruction error vector. 
Also, in this case, we can notice that the decoder can be represented as a shallow neural network having the dictionary elements as weights and the biases in terms of the reconstruction error vector (see Section \ref{subsec:dec_by_dictionary}). 

In the remainder of this section, we will describe two alternative ways (segmentation and dictionary learning) to obtain a decoder LRP method can be applied to, in more details. We experimentally tested our framework using both methods.

\subsection{Decoder by super-pixel segmentation}
\label{subsec:dec_by_segmentation}

Given an image $\mathbf{x}\in R^d$, we can obtain a partition of $\mathbf{x}$ composed of $m$ elements $P_h$ through any segmentation algorithm. 
We can associate to each element $P_h$  a vector $\mathbf{pv}^h \in R^d$ such that $ pv^h_i = 1$ if  $x_i \in P_h$, otherwise $pv^h_i = 0$. 
Thus, each element $P_h$ can be represented by the element-wise product between $\mathbf{x}$ and $\mathbf{pv}^h$, i.e., $\mathbf{v}^h = \mathbf{pv}^h \odot \mathbf{x}$, since this operation products selects all the pixels belonging to the element $P_h$. 

Consequently, we can decompose $\mathbf{x}$ as  $\mathbf{x}=\sum\limits_{h=1}^m u_h \mathbf{v}^h$, with $u_h=1$. 
Then, the decoder $D$ is a linear combination of the $\mathbf{v}^h$ with all the coefficients equal to $1$, which represent the encoding of the image $\mathbf{x}$ on the basis of the $m$ partition's elements.

Following  \cite{ribeiro2016}, in this paper we use the Quickshift segmentation algorithm \cite{vedaldi2008quick} where the elements of the partition are called super-pixels.

We assume that a possible explanation to the output of a given classifier can be obtained in terms of relevant super-pixels, where the relevance can be computed using an LRP-based procedure.

\subsection{Decoder by Sparse Dictionary Learning methods}
\label{subsec:dec_by_dictionary}
A sparse dictionary learning problem (see, for example, \cite{tessitore2011designing}) is a minimisation problem that one can  formally describe as follows.
\begin{equation}
\begin{split}
    \arg\min_{U,V} ||X -VU ||^2_F + \gamma_1 \sum\limits_{i=1}^k \Omega_V 
    (\mathbf{v}_i) \\
    \text{s.t. } \forall j, \Omega_U(\mathbf{u}_{i}) < 
    \gamma_2  
\end{split}
\label{sparseDictLearnProb}
\end{equation}
where $X \in R^{d \times n}$ is composed of $n$  experimental observations which are expressed as 
column vector $\mathbf{x}^i \in R^d$, $V$ is the dictionary, and the $k$ columns $\mathbf{v}^i$ of $V$ are the dictionary elements or atoms, subject to some sparsity constraint possibly.  
Each column of $X$ is approximated by a linear combination of the $k$ columns of $V$, subjects to some sparsity constraint potentially. 
Thus,  $U \in R^ {k \times n}$ is the matrix of the linear combination coefficients, i.e., the $i$-th column of $U$, $\mathbf{u}^i$, corresponds to the $k$ coefficients of the linear combination of the $k$ columns of $V$ to approximate $\mathbf{x}^i$, the $i$-th column of $X$.
$\Omega_V$  and $\Omega_U$ are some norms or quasi-norms that constrain or regularise the solutions of the minimisation problem, and $\gamma_1 \geq 0$ and $\gamma_2 \geq 0$ are parameters that control to what extent the dictionary and the coefficients are regularised. 

Elements of a dictionary can be used to compute explanations of a ML model response in terms of middle-level input features \cite{Apicella2019b,Apicella2019c,Apicella2019d,apicella2020middle}.

For the experiments presented in this paper, we obtain the dictionaries from a specific sparse dictionary learning method based on SSPCA\cite{jenatton2010sspca}. 
However, any dictionary learning/sparse coding method able to produce dictionaries that can be considered human-understandable can be used \cite{Apicella2019b}. 

Given a dictionary $V$ and an experimental observation $\mathbf{x}$ one can solve the minimisation problem as expressed in eq. \ref{sparseDictLearnProb}  with respect to the coefficients only, finding a single column vector $\mathbf{u}$. 
Consequently, $\tilde{\mathbf{x}}=V\mathbf{u}$ is an approximation of $\mathbf{x}$ with an error for each component equal to $\epsilon_h=x_h -\tilde{x}_h$.
Then, the decoder $D$ can be represented as a shallow neural network composed of just one weight layer $W$, $k$ input values and $d$ output neurons. 
Each output neuron $j$ has the identity as activation function and the bias equal to $\epsilon_j$. The weights associated to the connection going from the $j$-th input value to all the output neurons correspond to $j$-th $V$'s column. 
Consequently, given the decomposition of $\mathbf{x}$ as $V\mathbf{u}$ the decoder $D$ receives $\mathbf{u}$ as input and outputs $\mathbf{x}$. 

\section{Experimental assessment}
\label{sec:exp}

In this section, we describe the experiments performed and show the results obtained. We show a set of explanations produced by our approach using two different experimental setups. 

The former uses as middle-level features the super-pixel segmentation schema described in \ref{subsec:dec_by_segmentation}; the latter adopts the sparse dictionary approach described in \ref{subsec:dec_by_dictionary}. 
For the segmentation-based experiments, we used as classifier a VGG-16 \cite{Simonyan2015very} neural network trained on Imagenet, and as input images a subset of the STL-10 dataset \cite{coates2011analysis}.  
For the dictionary-based experiments, we use the JAFFE dataset \cite{lyons1998jaffe} and a custom neural network trained from scratch with a final accuracy of the $94\%$ on a test set. 
We chose to use a custom model because,  to the best of our knowledge, there are no reference models for this particular dataset in the current literature.  
Notice that for this type of middle-level features we used a more simple dataset since dictionary learning methods on large datasets can be very expensive in terms of computational costs.

We compare the results obtained by the proposed method (MLFR)  with two related methods proposed in the literature, LIME \cite{ribeiro2016} and  EM \cite{Apicella2019b,apicella2020middle}, and with a standard low-level feature method as LRP \cite{bach2015}. Notice that as we discussed in Section \ref{sec:related} the explanations returned by LIME and EM are based on features which can be considered of middle-level, but, differently from the MLFR approach,  they are built in a black-box approach relying on a proxy model instead of the actual model in case of LIME, and in terms of dictionary elements by a variant of the activation-maximisation method, in case of EM.
For the segmentation-based approach, we compared MLFR with LIME and LRP. For the dictionary-based approach, we compared our results with the ones produced by EM and LRP.
%Specifically, For the segmentation-based approach, we show results returned by LIME and LRP. The explanations returned by the former are based on features which can be considered of middle-level, but, differently from the MLFR approach, as discussed in Section \ref{sec:related}, they are built relying on a proxy model instead of the actual model (so, in a black-box approach). 
%LRP, instead, relies directly on the real model (so using a white-box approach) without using any proxy model but generating explanations as pixel-based heatmaps (low-level features). 
%For the dictionary-based approach, we compare our results with the ones produced by LRP and EM, a black-box approach which explains the ML model responses in terms of dictionary elements by a variant of the activation-maximisation method.
%We use different comparison algorithms: LIME and LRP for the segmentation-based approach on the STL-10 dataset, and EM for the dictionary-based approach on the JAFFE dataset, because they are known algorithms in the literature for those two types of features.
%makes explanations searching for a valid middle-level feature representations of the input which can explain the model behaviour. 
The segments and the dictionaries are obtained respectively using Quickshift \cite{vedaldi2008quick} (that is the same algorithm used by LIME to make the superpixel segmentation) and SSPCA \cite{jenatton2010sspca}. 

A visual comparison is not enough, and to give a quantitative evaluation of our results, we use the same strategy introduced in \cite{samek2016evaluating} and described in Section \ref{sec:quant_eval}.

\subsection{Qualitative results}
\begin{figure}
    \centering
    \hspace*{-.7in}
    \includegraphics[width=1.3\textwidth]{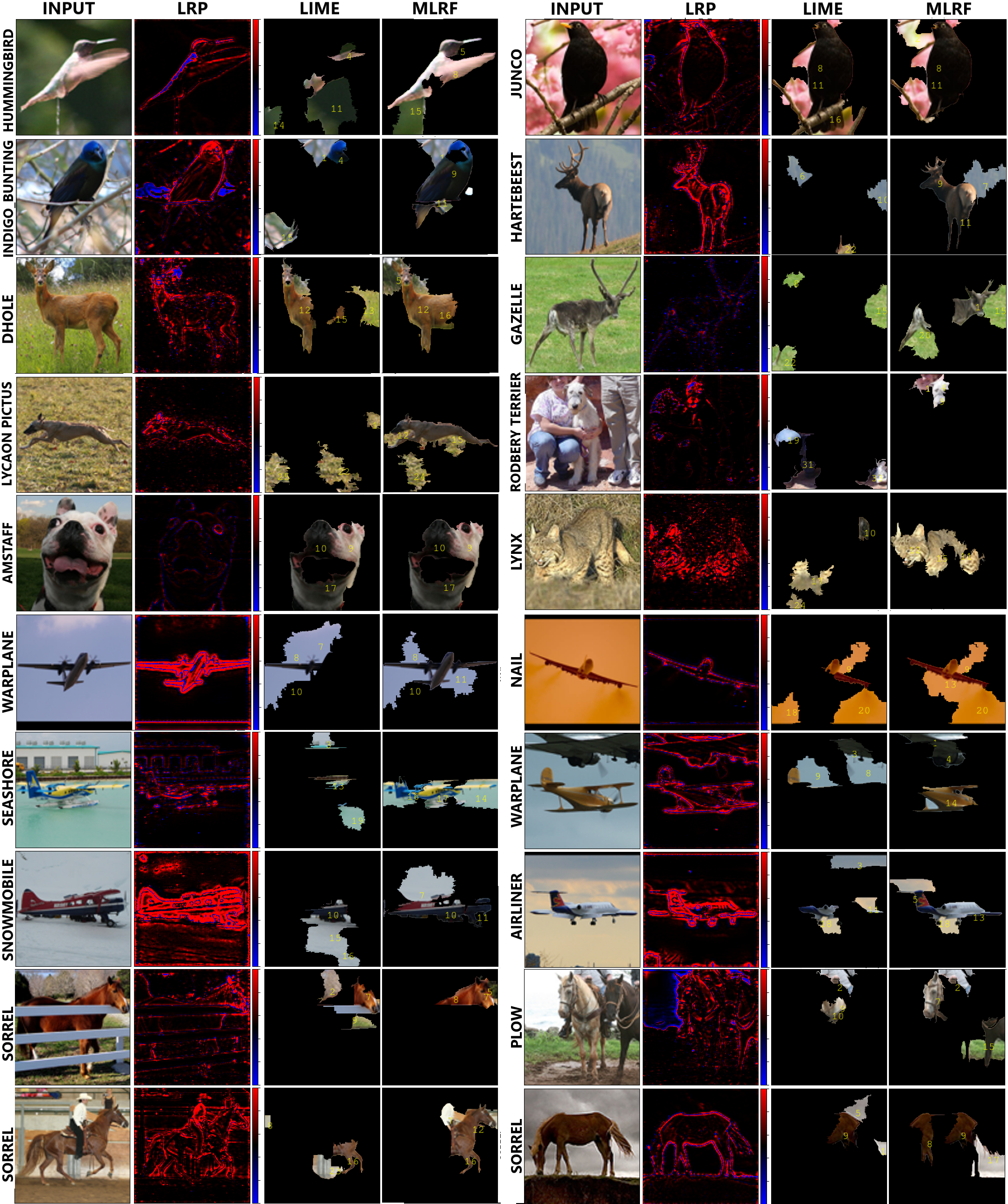}
    \caption{
    Results obtained from MLFR using image segmentation (section \ref{subsec:dec_by_segmentation}) on the STL10 dataset. For each input (1st and 5th column), we present the explanations produced by LRP method (2nd, 6th column) in terms of heatmaps (blue pixels indicate negative relevance, while red pixels indicate positive ones), LIME method (3th and 7th column) and MLFR (4rd and 8th column) as superimposition of the three superpixels with the highest relevance scores. The class returned by the classifier is reported for each input.}
    \label{fig_exp_ris1_stl10}
\end{figure}

\begin{figure}
    \centering
    \hspace*{-.8in}
    \includegraphics[width=1.3\textwidth]{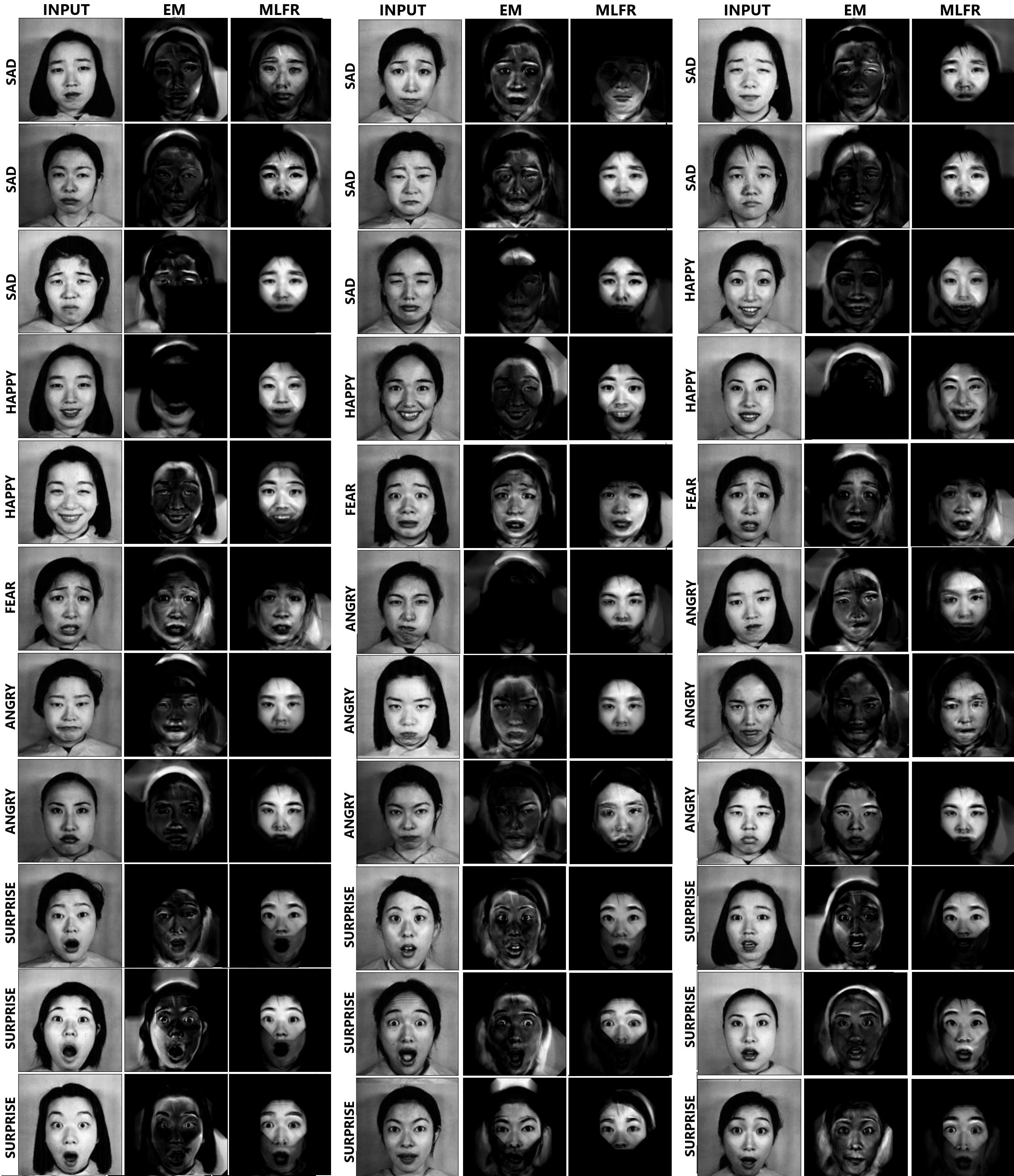}
    \caption{
    Results obtained from MLFR with sparse dictionaries (section \ref{subsec:dec_by_dictionary}) on the JAFFE dataset. For each input (1st, 4th and 7th column), we present the explanations obtained by EM method (2nd, 5th and 8th column) and MLFR (3rd, 6th and 9th column) as superimposition of the three atoms with the highest relevance scores. On the left of each input, we report the class returned by the classifier.}
    \label{fig_exp_ris2_jaffe}
\end{figure}

Some results of the two proposed strategies are shown in Figure \ref{fig_exp_ris1_stl10} for the superpixels-based approach and in Figure \ref{fig_exp_ris2_jaffe} for the dictionary-based approach. 
To make a comparison, we also report the explanations given by LIME and LRP methods for the superpixels-based approach and EM for the dictionary-based approach. 
For each input, we show the superposition of the three most relevant segments/atoms for LIME, EM and MLFR, and the heatmap produced by LRP. 
%\francesco{Non si capisce perché si usino metodi diversi su dataset diversi} (sal: inserito sopra)

With respect to LIME and EM, we can show that in several cases, the explanations produced by MLFR can be considered closer to what a human being expects from a classification system. 
For example, we expect that the output ``hummingbird'' and ``indigo bunting'' (figure \ref{fig_exp_ris1_stl10}, first and second row, first column) is due mainly by the presence of the main components of a bird in the image, neglecting non-relevant part as background sprigs. 
Similar considerations can be done for the ``hartebeest'' and the ``gazelle'' (figure \ref{fig_exp_ris1_stl10}, second and third row, fifth column) and several other inputs shown in the figure. In particular, the ``Bedlington Terrier'' input (figure \ref{fig_exp_ris1_stl10}, fourth row, fifth column) provides an interesting case due to the presence of several hypothetical relevant candidates (the several human being parts) which can lead the classifier toward different classification. 
The proposed method, in agreement with LRP, highlights that the dog face is one of the main discriminative parts behind the classifier's choice. 
The different results returned by LIME can be due to several factors, such as a sub-optimal training procedure of the proxy model or the inadequacy of the proxy model in representing the real one.

Similar consideration can be done for the results shown in figure \ref{fig_exp_ris2_jaffe} inherent the dictionary-based approach.
We show inputs for several classes of the Jaffe dataset (SAD, SURPRISE, HAPPY, FEAR, ANGRY) and the results obtained respectively by the EM method and the proposed MLFR. 
It is possible to see how the proposed method highlights atoms which better characterise the faces, as we would expect by an emotion classifier.  
MLFR highlights details concerning facial expressions as the open mouth and the eyes for the inputs classified as ``SURPRISE'' or the smiling expression for the input classified for ``HAPPY'' (for example, on the figure  \ref{fig_exp_ris2_jaffe} see the results of the inputs on the first, fourth and seventh columns of the 5th and 8th row). 
The results produced by EM method, instead, seems more confused and less clear and intuitive. 
As far as the EM method is not based on proxy models, it is again a black-box approach based only on the input/output relations of the classifier, so without any knowledge to the real inner state of the model.
Furthermore, EM needs several hyperparameters to be set, which can affect the reliability of the results produced. 

Notice that for reasons of space we do not report the results obtained by LRP, since the qualitive comparison is similar to the one for the STL10 dataset.
%Interestingly, the explanations produced by LIME for the VGG16 and MobileNet are very similar for the same input, while the results given by LRP and our proposed approach results to be different in several cases. For example, the classification of the red bird in figure  as ``robin`` and ``goldfinch''  by VGG and MobileNet respectively is ``explained'' by LIME by the body area of the bird. Instead, the proposed approach highlights that the oracle answer was probably conditioned by the bird's head and the paws. The ``sorrel'' class assigned to the horse image input is explained by LIME as returning mainly background elements, unlike LRP which already gives us an indication that the area involved in the decision can be the nose. This suggestion is confirmed by our proposed method which highlights the nose's zone as relevant both in VGG and MobileNet.
%Also in the fire-engine case the LIME explanations seem to be similar between them, but the proposed method showed us that a discriminant area was, at least in one case (MobileNet), the fire-track ladder. In the deer image, both LIME explanations do not consider the horns, which could be a reasonable discriminant factor, despite the presence in both the explanations produced by our approach.
%Also in the aircraft carrier, despite the similar LIME explainations, our strategies suggest that the motivation behind the classifications could be very different.

\subsection{Quantitative evaluation}

\label{sec:quant_eval}
\begin{figure}[t]
     \centering
     \begin{subfigure}[b]{0.45\textwidth}
       \centering
    \includegraphics[width=\textwidth]{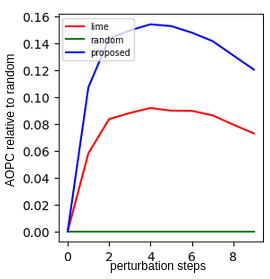}
    \caption{AOPC curve of the proposed method (segmentation-based, section \ref{subsec:dec_by_segmentation}) compared with the LIME method.}
    \label{fig_AOPC1}
     \end{subfigure}
     \hfill
     \begin{subfigure}[b]{0.45\textwidth}
    \centering
    \includegraphics[width=\textwidth]{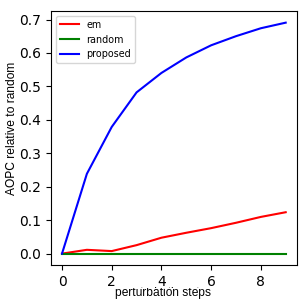}
    \caption{AOPC curve of the proposed method (DL-based, section \ref{subsec:dec_by_dictionary}) compared with the EM method.}
    \label{fig_AOPC2}
     \end{subfigure}
     \hfill
     \caption{Comparison of the AOPC curves of the methods used in the experiments. As made in \cite{samek2016evaluating}, all the curves have been plotted relatively to a random AOPC curve, which was obtained following a random order instead of a relevance order during the image perturbation steps.}
\end{figure}

In the previous section, we show the explanations obtained in terms of the most relevant middle-level features selected by MLFR compared against the ones selected by some related works proposed in the literature. 
However, all the consideration we made are based only on subjective evaluations, and an objective and quantitative evaluation of the explanation methods is still an open research problem.

A possible quantitative evaluation framework was proposed in  \cite{samek2016evaluating} with \textit{region flipping}, a generalisation of the \textit{pixel-flipping} measure proposed in \cite{bach2015}.  
In a nutshell, given an image classification to explain, regions of a given size are substituted  iteratively, following the descending relevance order assigned to the central pixel (MoRF, Most Relevant First) by the explanation method. At each step, the difference between the original class score returned by the model and the score returned on the perturbed input is computed, generating a curve (MoRF curve). 
We expect that the better the explanation method is, the stronger the difference between the scores is. Repeating this process for several images and averaging between them, it is possible to obtain the \textit{Area Over the MoRF Perturbation Curve} (AOPC):
$$AOPC=\frac{1}{L+1}<\sum\limits_{k=0}^L f(x^{(0)})-f(x^{(k)})>_{p(x)}$$
where $<\cdot>_{p(x)}$ is the average over the dataset images, $L$ is the number of regions and $x^{(k)}$ is the input at $k-$th perturbation step. 
If the regions are well-ranked (so, relevant regions have a higher relevance), we expect that the resulting AOPC values are large, so we can infer that the largest the AOPC value is, the better the explanation method is.
The original region-flipping method was originally defined for pixel-based heatmaps using regions of fixed size ($9\times 9$ in \cite{samek2016evaluating}). 
However, it is easily adapted to our proposed method and LIME, considering that each middle-level feature is a single region. As a perturbation scheme, we adopt the same used in \cite{samek2016evaluating}, changing each pixel in the region with a value sampled from the Uniform distribution. 
In figure \ref{fig_AOPC1} we plot the AOPC curve for LIME and our proposed method on the VGG16 model, showing that MLFR outperforms LIME in terms of AOPC curve, suggesting that the former, on average, gives a more reliably relevance score respect to the latter. We hypothesise that LIME, exploiting a proxy classifier which \textit{emulates} the real one, may not capture the real ``reasons'' behind the choices made by a classifier, so assigning scores to the features in a manner which not reflect the real inner state of the classifier. 
Similar results are shown in figure \ref{fig_AOPC2}, where the results of the proposed approach are compared with the EM method. 
Again, in this case, the proposed method shows better results in terms of AOPC values, giving better reliability to the explanations produced.
\section{Conclusions}
\label{sec:conclusion}
In this work, we propose MLFR, a novel XAI method based on middle-level features. 
The proposed method generalises the well-known LRP method, initially proposed for low-level features (such as pixels for image domain), to middle-level features, returning data representations which can be interpreted by a human. 
We describe how the proposed method can be easily adapted to several classes of middle-level features. 
For instance, we show how two different middle-level input representations can be suitable for the proposed method, the former based on image segments directly obtained from the input to explain, the latter on a more general set of elements which can be constructed through some dictionary learning approach. However, nothing prevents to use  other representations. 

To evaluate the proposed method, we adapt the quantitative measure described in \cite{samek2016evaluating}, proposed initially for pixelwise-based methods, to middle-level feature methods, and we make a comparison with others middle-level features approaches present in literature. The results of the experiments that we carried out are encouraging, both under the qualitative point of view, giving explanations that can be easily interpretable by the human being, and the quantitative point of view, giving performances in terms of AOPC curve which are comparable to other methods present in the current literature.
\section*{Acknowledgments}
The research presented in this paper was partially supported by the national project Perception, Performativity and Cognitive Sciences (PRIN Bando 2015, cod. 2015TM24JS 009).

%%%%%%%%%%%%%%%%%%%%%%%%%%%%%
 \bibliographystyle{splncs04}
 \bibliography{bibliografia}
% link homepage: https://edl-ai-icpr.labri.fr/
\end{document}